\documentclass{article}
\usepackage{graphicx}
\usepackage{fullpage}
\usepackage[bottom, flushmargin]{footmisc}
\usepackage{natbib}
\usepackage{booktabs}
\usepackage{setspace}
\usepackage{titlesec}
\usepackage{subcaption}
\usepackage{pdfpages}

\titleformat{\section}{\normalfont\fontsize{12}{0}\bfseries}{\thesection}{1em}{}
\titlespacing\section{0pt}{0pt}{0pt}

\title{Memory Is All You Need: Testing How Model Memory \\  Affects LLM Performance in Annotation Tasks}  \vspace{1cm}

\author{\hspace{-1cm} 
Joan C. Timoneda\thanks{Assistant Professor, Department of Political Science, Purdue University. E-mail: timoneda@purdue.edu} 
\and
Sebastián Vallejo Vera\thanks{Assistant Professor, Department of Political Science, University of Western Ontario. E-mail: sebastian.vallejo@uwo.ca}}

\date{}

\begin{document}

\maketitle

\begin{abstract}

\noindent Generative Large Language Models (LLMs) have shown promising results in text annotation using zero-shot and few-shot learning. Yet these approaches do not allow the model to retain information from previous annotations, making each response independent from the preceding ones. This raises the question of whether model memory---the LLM having knowledge about its own previous annotations in the same task---affects performance. In this article, using OpenAI's GPT-4o and Meta's Llama 3.1 on two political science datasets, we demonstrate that allowing the model to retain information about its own previous classifications yields significant performance improvements: between 5 and 25\% when compared to zero-shot and few-shot learning. Moreover, memory reinforcement, a novel approach we propose that combines model memory and reinforcement learning, yields additional performance gains in three out of our four tests. These findings have important implications for applied researchers looking to improve performance and efficiency in LLM annotation tasks.

\end{abstract}

\thispagestyle{empty}

\pagebreak

\setcounter{page}{1}
\setstretch{2}

\section*{Introduction}

\noindent Recent work in the social sciences explores the benefits and limitations of using Generative Large Language Models (LLMs) as text annotators, with promising results. LLMs often outperform human coders \citep{gilardi2023chatgpt} and provide comparable accuracy when labeling political text \citep{overos2024coding, ornstein2023train}, even across multiple languages \citep{heseltine2024large}. Most of these studies craft their LLM prompts carefully, using zero-shot or few-shot learning with chain of thought (CoT) reasoning \citep{wei2022chain,chen2023you}. However, there is one dimension to LLM prompting that remains understudied: how does \textit{model memory} affect performance in annotation tasks? While few-shot learning provides some examples to the LLM to improve its performance, it does not allow the model to learn from its own previous task-specific classifications. Doing so, we find, significantly improves performance over zero-shot and few-shot learning with CoT.

\begin{table}[!b]
    \centering
    \scalebox{0.8}{\begin{tabular}{cc}
    Article & Prompting Strategy \\
    \midrule
    \cite{ornstein2023train} & Few-Shot + \textbf{No memory} \\
    \cite{alizadeh2023open} & Few-Shot + \textbf{No memory} + CoT \\
    \cite{mellon2024ais} & Few-Shot (25) + \textbf{No memory} \\
    \cite{gilardi2023chatgpt} & Zero-Shot + \textbf{No memory} \\
    \cite{kuzman2023chatgptbeginningendmanual} & Zero-Shot + \textbf{No memory} \\
    \cite{Huang_2023} & Zero-Shot + \textbf{No memory}  \\
    \cite{tornberg2023chatgpt} & Zero-Shot + \textbf{No memory} \\
    \cite{nguyen2024human} & Zero-Shot + \textbf{No memory}  \\
    \cite{laskar2023cqsumdp} & Zero-Shot + \textbf{No memory} \\
    \cite{ostyakova2023chatgpt} & Zero-Shot + \textbf{No memory} + Multi-Step\\
    \cite{reiss2023testing} & Zero-Shot + \textbf{No memory}  \\
    \cite{le2023uncovering} & Zero-Shot + \textbf{No memory}  \\
    \cite{muller2024nostalgia} & Zero-Shot + \textbf{No memory}  \\
    \cite{steinert2024user} & Zero-Shot + \textbf{No memory} \\
    \cite{rathje2024gpt} & Zero-Shot + \textbf{No memory}  \\
    \cite{lee2024people} & Zero-Shot + \textbf{No memory}  \\
    \cite{dickson2025effects} & Zero-Shot + \textbf{No memory} \\
    \cite{heseltine2024large} & Zero-Shot + \textbf{No memory} \\
    \midrule
\end{tabular}}
    \caption{Prompting Strategies of Articles using LLMs as Annotators}
    \label{tab:lit}
\end{table}

A brief survey of extant research reveals the novelty of our memory-based approach. Table \ref{tab:lit} lists publications that use LLMs as annotators alongside their prompting strategy. It shows that most of the focus has been on the prompt and its design. Many authors use zero-shot learning, where the LLM is given a prompt without any examples to guide the annotation. Some use few-shot learning, where the LLM receives a small set of examples in the prompt to illustrate how to annotate new text \citep{alizadeh2023open,mellon2024ais, ornstein2023train}. The highest-performing few-shot approaches use chain-of-thought prompting, `where large LLMs are provided with both the question and a step-by-step reasoning answer as examples' \citep{alizadeh2023open}.\footnote{Another approach is multi-step approach, where researchers use a tree-like annotation scheme that breaks labeling tasks into multiple steps with simpler questions that build up to the desired annotation task \citep{ostyakova2023chatgpt}.} Yet one aspect remains consistent across these works: authors use a no-memory strategy, sending each new text to the LLM as an independent user message. None of the approaches reviewed allow the model to retain previous information, gain knowledge and improve its performance --as human annotators do. Furthermore, few-shot learning techniques only provide \textit{correct} answers to the models, rather than allowing the model to understand both successes and failures.\footnote{\citet{chia2023contrastive} propose \textit{contrastive CoT prompting}, where they add examples with incorrect and correct explanations to the CoT prompt in order to show the LLM how not to reason. They find improvements in arithmetic reasoning and factual QA. We think this logic supports our memory-based approaches, which take all classifications regardless of accuracy.}

In this article, we analyze how model memory affects annotation performance by testing four different LLM \textit{interaction approaches}: (1) zero-shot learning with no memory, (2) few-short learning (CoT) with no memory, (3) memory prompting and (4) memory reinforcement. The first two approaches reflect common state-of-the-art practices in the literature. The latter two are novel approaches we introduce in this article that, we find, lead to significant performance gains in applied political science tasks. For (1), we follow the literature and provide a system prompt\footnote{System prompts are parts of the overall LLM prompt used to provide general instructions for a task.} without any examples, and submit each new classification request as a user message independent from previous ones. We give no examples to the LLM. Importantly, the model keeps no explicit memory of the task at hand.\footnote{While generative LLMs do not store previous text instances in an API session, the model session itself remains active through the annotation task. OpenAI has stated that GPT models do not retain memory between API calls \citep{achiam2023gpt}. Since no textual information is kept, the model cannot improve its classification based on patterns it identifies in previous annotations and behavior. However, there is a chance that processing optimizations within model may subtly influence subsequent queries.} For (2), the current state-of-the-art approach, we apply few-shot learning with CoT reasoning by adding 10 examples to the system prompt in (1).\footnote{The number of examples for few-shot learning ranges from 1 to 25, but a majority lie within the 1-10 range.} In separate sessions, we ask the LLM to classify each example and provide a reasoning. We then add the text, the true classification, and the full reasoning to the few-shot learning with CoT prompt.\footnote{See \citep{alizadeh2023open} for further details on applying few-shot learning with CoT.} 

For (3), our memory prompting approach, the model keeps a memory of previous text instances it has classified before it classifies a new one. We accomplish this by sending the conversation history to the model for every new annotation request. The model thus has explicit knowledge of previous texts and how it classified them, and can use this knowledge to classify new text instances -- as a human annotator would. It does not, however, know if its previous annotations were correct or not.\footnote{Note that in the tests below we send only a partial history with the most recent 200 classifications to avoid API rate limit errors and to limit costs for applied researchers. While sending the full conversation history would be ideal, our tests (especially with GPT-4o) showed that API errors increased significantly above 200 previous classifications. If texts are long, errors could emerge sooner, so the researcher may need to adjust this number downward to fit their task. Moreover, while GPT-4o is not an expensive model, costs can quickly accumulate when passing long conversation histories to the model for every single classification request.} The logic behind this approach is that the LLM may improve its performance over time as it gains more information on the task at hand.

Finally, memory reinforcement --(4)-- is also a new method we propose that is similar in logic to reinforcement learning \citep{ouyang2022training}. This approach requires a ground truth annotated training set whose observations are sent sequentially to the LLM for classification. We set aside 20 percent of observations in the dataset for training, similar to the common 20/80 percent train-test split in machine learning. After each training annotation, the researcher informs the model whether it was correct or not, sending a punishment or reward message. We record the interaction, update the history of interactions, and send it back to the model together with the next textual observation to annotate. After classifying the full set of training examples, the model is asked to annotate the rest of the sample with knowledge of the full interaction during training. As with the second approach, the model always has a memory of previous performance. Since reinforcement learning is part of LLM model training, we hypothesize that this approach should yield the best performance.\footnote{OpenAI's GPT-4, for instance, is fine-tuned using Reinforcement Learning from Human Feedback, which is especially relevant in filtering answers to make them more helpful and less harmful and biased \citep{achiam2023gpt}.} By way of an example, if our dataset has 600 observations, we set aside 120 (20\%) as the ground truth training set. We generate a system prompt for classification and, after sending each text for annotation, we respond with a message rewarding the model if correct, and punishing it if not, together with the correct answer.\footnote{In Appendix A we provide the reward and punishment prompts used in our examples.} We save the LLM's classification, our response to it, and its response. When we submit the request to classify the second observation in the training set, we send the saved data plus the new text to classify (the system prompt is always at the top with the task instructions). We follow this procedure with each new text from the training set. Once the full set of 120 has been classified, we save that full conversation history, and submit it together with every new text instance for the rest of the unlabeled sample (480 observations, or 80\%).

To test the performance of these four strategies, we use state-of-the-art versions of OpenAI's and Meta's generative models at the time of writing -- GPT-4o and Llama 3.1. We apply these models to two political science classification tasks, political nostalgia in party manifestos \citep{muller2024nostalgia} and incivility on Twitter \citep{rheault2019}. Our main result and contribution is to show that memory matters: the memory prompting and memory reinforcement approaches significantly improve performance in annotation tasks, using both GPT-4o and Llama 3.1. This finding has major implications for applied researchers seeking to use generative LLMs in annotation tasks. Improved performance translates into better annotations and higher-quality data, while a reduced need for few-shot examples puts less emphasis on the prompt and improves efficiency.

%, which may in turn benefit replicability.\footnote{Memory approaches, while they send the conversation history to the model, are by definition zero-shot, so researchers need to devote considerable time and effort to example selection for the few-shot prompt, improving time efficiency. Because the prompt is partially less reliant on a researcher's example choices, memory approaches should help with replicability over multiple runs.} %% Replication implication: by making prompts less important, and memory more important, you rely less on the prompt which should aid both consistency in the results and better replications.

\vspace{1cm}

\section*{Data and Methods}

\noindent Our first set of tests uses data on political nostalgia from \cite{muller2024nostalgia}. The authors coded specific sentences from party manifestos in 24 European countries into a binary measure. Four human annotators coded the data into either nostalgic or not nostalgic,\footnote{The authors define nostalgia as a `predominately positive emotion that is associated with recalling memories of important or momentous events, usually experienced with close others' \citep{muller2024nostalgia}.} and the authors aggregated these into a variable where a sentence is classified as nostalgic if three of the four coders agreed.\footnote{The authors translated each non-English sentence into English before classification. They also used different dictionary and machine learning approaches for classification, but we will use the human-annotated data for our purposes.} The dataset contains 1,200 human-annotated sentences, with 151 coded as nostalgic and 1,049 as not nostalgic. To maintain consistency in the comparison between our two samples, and considering the significant imbalance of these data, we created a subsample of 600 observations, keeping all 151 nostalgic observations and drawing a random sample of 449 non-nostalgic ones. Sentences from the sample were 20.83 words long on average, with a minimum of 3 and a maximum of 174 words, similar to the original sample. 

Our second set of tests uses \citeauthor{rheault2019}'s (\citeyear{rheault2019}) data on incivility on Twitter. The authors collected 2.2 million tweets about US and Canadian public officials between April and July, 2017, and manually coded a sample of 10,000 tweets as civil or uncivil through crowd-sourcing. A tweet was deemed uncivil if it contained ``(1) swear words; (2) vulgarities; (3) insults; (4) threats; (5) personal attacks on someone's private life; or (6) attacks targeted at groups (hate speech)'' \citep{rheault2019}. For our tests, we drew a balanced random sample of 600 tweets from the authors' training data, with 300 civil and 300 uncivil tweets.\footnote{We decided to draw a sample that had a sufficiently large N to maintain power but also would keep costs down, especially when using GPT-4o.} Texts were 16.76 words long on average, with a minimum of 3 and a maximum of 31 words, all similar to the original sample.  We selected this sample to provide a test with a different type of text data, one with less formal and less specialized political language than data from political party manifestos.

For each of these, we use two LLMs and compare their performance -- GPT-4o and Llama 3.1-70B, the latest releases by OpenAI and Meta.\footnote{We focus on two widely available models that are relatively easy to implement through cloud-based servers. GPT-4o can be implemented on a pay-per-token basis using OpenAI's API. Meta's Llama family is open-source and free, an important element when considering our models of choice \citep{palmer2024using}. Note also that ChatGPT is the publicly available version of OpenAI's models, and uses GPT-4o in its `pro' version.} GPT-4o is an LLM trained on a vast corpus to produce text based on a prompt by the user. It is faster and more efficient than GPT-4 while retaining similar performance.\footnote{GPT-4o is also significantly cheaper than GPT-4, which is one of the reasons we selected it over GPT-4 considering its similar performance (See Table B1 in Appendix B for more details on costs).} These models are then optimized for dialogue using Reinforcement Learning with Human Feedback \citep{christiano2017deep}, where the model produces multiple answers from a single prompt, and a human evaluates the best response. This response is then used as a reward function to improve model's output \citep{achiam2023gpt}. Similar to OpenAI's GPT LLMs, models from the Llama family are fine-tuned and then improved by combining techniques that use preference ranking to signal to the model the best response from several LLM responses \citep{wu2024meta}. Given these learning mechanisms, we expect that retaining previous iterations of the annotation task, even if correct labels remain unknown, should improve overall model performance.    

For both LLMs, we run the following tests on each of the two datasets. First, we ask the LLMs to annotate the data \textit{without memory}, i.e. we submit each new classification request as an independent user message, with the model storing no memory of previous interactions. We do this for both the zero-shot and few-shot CoT approaches. We classify the entire dataset 10 times and calculate and report the average performance over all the runs. Second, we follow \textit{memory prompting}, keeping the most recent 200 annotations in the conversation history when submitting a new classification request. We also generate 10 different classification runs. Lastly, we apply \textit{memory reinforcement} over 10 runs.\footnote{In Appendix C, we also provide results for both the memory prompting and memory reinforcement approaches using the few-shot CoT prompt. Adding the few-shot examples does not yield additional performance benefits for these two approaches.} For each of these tests, we use the OpenAI library in Python to query both LLMs and obtain an annotation.\footnote{For Llama, we use the OpenAI library and code paired with a server that hosts the Llama model. The code is available in the replication materials.} 

For the political nostalgia tests, the system prompt is: ``Please classify the following sentence as containing `nostalgic' or `not nostalgic' ideas or subtexts about politics, society, or the past. [New line] Return only the name of the category.'' For the incivility classification task, we use the following base system prompt: ``Please classify the following sentence as `civil' or `incivil'. [New line] Return only the name of the category.'' Each sentence is then submitted to the LLM as a new user message. For the few-shot CoT approach, we add ten examples to the prompt (i.e., text, answer, and reasoning), and then ask the model to classify each new text into one of two categories.\footnote{Each example sentence is indicated with `text:'; each correct classification with `Answer:', and each reasoning text with `Reasoning:'. All these elements are always separated by a new line.} For memory prompting, we save the conversation history after every annotation and run the model again asking it to classify a new text, keeping a set of previous annotations in the conversation history.\footnote{That is, first we append the previous classification to the conversation history. Then, we append a new text to classify to the conversation history as a new user prompt. Lastly, we make a new request to the model passing the conversation history, which includes both previous interactions plus the latest text to classify. When interacting with the API this way, the researcher can have a conversation with the model through the API. The code to implement this approach is in the replication materials.} For \textit{memory reinforcement}, we follow the same procedure as with memory prompting with the hold-out training set, but add a reinforcement step where the model is told whether it was the correct or incorrect classification (see Appendix A for the reward and punishment messages). We also save the reinforcement interaction and submit it to the model before it classifies the next text. We use a temperature of 0.7 across all models to maintain consistency. We keep the rest of the hyperparameters at their default values. 

Each test results in a set of classifications that we compare with the human-annotated ground truth to evaluate model performance across the various prompting approaches. In the results section, we report three key performance metrics: $F1$ scores, precision and recall, following \citet{timoneda2025behind, timoneda2025}. $F1$ is the harmonic mean between precision and recall and is a better metric than accuracy in unbalanced samples. While our incivility sample is balanced (300 civil and 300 uncivil), the models often produced unbalanced classifications, in which case $F1$ is also the best metric for true model accuracy. We report precision and recall to understand the extent of the imbalance between false positives and false negatives, which will be an important part of our results discussion below. All three statistics range between 0 and 1.

\vspace{1cm}

\section*{Results: Nostalgia}

\noindent The results for the nostalgia dataset are in Table 2, and generally show larger differences between current no-memory approaches and our two memory-based alternatives. Table 2a displays the results for GPT-4o. Again, as with incivility, few-shot CoT bests zero-shot learning (both without memory) by 12.66 percent on average when comparing $F1$ scores. The improvement is 33.2 percent with Llama. These results confirm the well-known finding in the literature that few-shot CoT significantly improves on zero-shot performance. More importantly, however, both memory prompting and memory reinforcement show significant improvements. With GPT-4o, memory prompting yields performance gains of 5.57 percent on average when compared to few-shot CoT with no memory. Especially relevant is the gain in the nostalgic category, which goes from 0.673 to 0.753 with memory prompting, an 11.23 percent improvement. With Llama, the gain is 9.08 percent on average, with the nostalgic category's $F1$ score also improving substantially from 0.626 to 0.755, or 20.61 percent. All the $F1$ score differences between few-shot CoT and memory prompting are statistically significant at the 0.01 level in a two-sample t-test.

\begin{table}[!b]

\begin{subtable}{1\textwidth}
    \centering
    \caption{GPT-4o}
    \scalebox{0.75}{\def\arraystretch{1.2}
\begin{tabular}{l||c|c|c|c||c|c|c|c||c|c|c|c||c|c|c|c}
\toprule 
\multicolumn{1}{c||}{} & \multicolumn{4}{c||}{Zero-Shot No-Memory} & \multicolumn{4}{c||}{Few Shot No-Memory (CoT)} & \multicolumn{4}{c||}{Memory Prompting} & \multicolumn{4}{c}{Memory Reinforcement} \\
\midrule 
          & Prec. & Rec. & $F1$ & $SD_{F1}$ & Prec. & Rec. & $F1$ & $SD_{F1}$   & Prec. & Rec. & $F1$ & $SD_{F1}$   & Prec. & Rec. & $F1$ & $SD_{F1}$    \\
\midrule
Not Nostalgic & 0.825 & 0.943 & 0.880 & 0.004 & 0.890 & 0.894 & 0.892 & 0.004 & 0.934 & 0.885 & 0.908 & 0.008 & 0.981 & 0.863 & 
 \bf{0.918} & 0.008 \\
Nostalgic     & 0.707 & 0.407 & 0.516 & 0.015 & 0.681 & 0.673 & 0.677 & 0.011 & 0.704 & 0.812 & 0.753 & 0.024 & 0.700 & 0.948 & \bf{0.804} & 0.013 \\
Overall       & 0.796 & 0.808 & 0.789 & 0.006 & 0.837 & 0.838 & 0.838 & 0.006 & 0.876 & 0.867 & 0.869 & 0.012 & 0.910 & 0.884 & 
 \bf{0.889} & 0.009 \\
\bottomrule
\end{tabular}
}
    \label{tab:gpt_nost}
\end{subtable}
\vspace{0cm}

\begin{subtable}{1\textwidth}
    \centering
    \caption{Llama 3.1}
    \scalebox{0.75}{\def\arraystretch{1.2}
\begin{tabular}{l||c|c|c|c||c|c|c|c||c|c|c|c||c|c|c|c}
\toprule 
\multicolumn{1}{c||}{} & \multicolumn{4}{c||}{Zero-Shot No-Memory} & \multicolumn{4}{c||}{Few Shot No-Memory (CoT)} & \multicolumn{4}{c||}{Memory Prompting} & \multicolumn{4}{c}{Memory Reinforcement} \\
\midrule 
          & Prec. & Rec. & $F1$ & $SD_{F1}$  & Prec. & Rec. & $F1$ & $SD_{F1}$  & Prec. & Rec.   & $F1$ & $SD_{F1}$ & Prec. & Rec. & $F1$ & $SD_{F1}$           \\
\midrule
Not Nostalgic   & 0.787 & 0.975 & 0.871 & 0.002 & 0.869 & 0.892 & 0.881 & 0.002 & 0.961 & 0.837 & 0.894 & 0.016 & 0.998 & 0.831 & \bf{0.903} & 0.010 \\
Nostalgic       & 0.741 & 0.215 & 0.334 & 0.011 & 0.653 & 0.602 & 0.626 & 0.009 & 0.655 & 0.897 & 0.755 & 0.022 & 0.658 & 0.969 & 
 \bf{0.783} & 0.015 \\
Overall         & 0.776 & 0.784 & 0.736 & 0.003 & 0.815 & 0.819 & 0.817 & 0.004 & 0.884 & 0.853 & 0.859 & 0.017 & 0.905 & 0.866 & 
 \bf{0.873} & 0.011 \\
\bottomrule
\end{tabular}
}
    \label{tab:llama_nost}
\end{subtable}

\caption{Nostalgia results for the three different interaction approaches, by LLM. Performance metrics averaged over 10 runs.}
\end{table}

The results for the memory reinforcement approach are even stronger. Reinforcement yields an average performance improvement of 9.26 percent with GPT-4o and 11.48 percent with Llama when compared to few-shot CoT. Performance gains in the nostalgic category are especially noteworthy, as the reinforcement $F1$ score increases from 0.677 in few-shot CoT with GPT-4o to 0.804 with reinforcement, or 18.76 percent. With Llama, these numbers are 0.626 and 0.783 respectively, or a difference of 25.08 percent. Again, all these differences are statistically significant at the 0.01 level in a two-sample t-test. 

\vspace{1cm}

\section*{Results: Incivility}

\begin{table}[!b]

\begin{subtable}{1\textwidth}
    \centering
    \caption{GPT-4o}
    \scalebox{0.77}{\def\arraystretch{1.2}
\begin{tabular}{l||c|c|c|c||c|c|c|c||c|c|c|c||c|c|c|c}
\toprule 
\multicolumn{1}{c||}{} & \multicolumn{4}{c||}{Zero-Shot No-Memory} & \multicolumn{4}{c||}{Few Shot No-Memory (CoT)} & \multicolumn{4}{c||}{Memory Prompting} & \multicolumn{4}{c}{Memory Reinforcement} \\
\midrule 
          & Prec. & Rec. & $F1$ & $SD_{F1}$   &
          Prec. & Rec. & $F1$ & $SD_{F1}$   & Prec. & Rec.   & $F1$       & $SD_{F1}$ & Prec. & Rec. & $F1$ & $SD_{F1}$          \\
\midrule
Civil   & 0.935 & 0.627 & 0.751 & 0.015 & 0.943 & 0.680  & 0.790 & 0.004 & 0.882 & 0.766 & \bf{0.820} & 0.008 & 0.923 & 0.723 & 0.810 & 0.012 \\
Uncivil & 0.720 & 0.960 & 0.822 & 0.007 & 0.750 & 0.960 & 0.842 & 0.003 & 0.793 & 0.897 & 0.842 & 0.005 & 0.772 & 0.939 & \bf{0.848} & 0.007\\
Overall & 0.828 & 0.792 & 0.786 & 0.011 & 0.846 & 0.820 & 0.816 & 0.004 & 0.838 & 0.832 & \bf{0.831} & 0.006 & 0.848 & 0.831 & 0.829 & 0.010\\
\bottomrule
\end{tabular}
}
    \label{tab:gpt}
\end{subtable}
\vspace{0cm}

\begin{subtable}{1\textwidth}
    \centering
    \caption{Llama 3.1}
    \scalebox{0.77}{\def\arraystretch{1.2}
\begin{tabular}{l||c|c|c|c||c|c|c|c||c|c|c|c||c|c|c|c}
\toprule 
\multicolumn{1}{c||}{} & \multicolumn{4}{c||}{Zero-Shot No-Memory} & \multicolumn{4}{c||}{Few Shot No-Memory (CoT)} & \multicolumn{4}{c||}{Memory Prompting} & \multicolumn{4}{c}{Memory Reinforcement} \\
\midrule 
          & Prec. & Rec. & $F1$ & $SD_{F1}$  & Prec. & Rec. & $F1$ & $SD_{F1}$  & Prec. & Rec.   & $F1$ & $SD_{F1}$ & Prec. & Rec. & $F1$ & $SD_{F1}$           \\
\midrule
Civil   & 0.957 & 0.480 & 0.639 & 0.007 & 0.947 & 0.627 & 0.755 & 0.005 & 0.874 & 0.722 & 0.795 &  0.012 & 0.909 & 0.763 & \bf{0.830} & 0.013\\
Uncivil & 0.653 & 0.978 & 0.783 & 0.003 & 0.724 &  0.966 & 0.828 & 0.003 & 0.764 & 0.896 & 0.828 & 0.005 & 0.797 & 0.920 & \bf{0.855} & 0.007\\
Overall & 0.805 & 0.729 & 0.711 & 0.005 & 0.835 & 0.798 & 0.792 & 0.004 &  0.819 & 0.809 & 0.811 & 0.008 & 0.853 & 0.843 & \bf{0.842} & 0.010\\
\bottomrule
\end{tabular}
}
    \label{tab:llama}
\end{subtable}

\caption{Incivility results for the three different interaction approaches, by LLM. Performance metrics averaged over 10 runs.}
\end{table}

\noindent Table 3 reports the main results on the incivility dataset for GPT-4o (a) and Llama 3.1 (b). Comparing across prompting strategies, performance is lowest for the zero-shot no-memory approach for both LLMs. With GPT-4o, this method yields an $F1$ score of 0.751 for civil tweets, 0.822 for uncivil tweets, and 0.786 overall. As expected, few-shot CoT with no memory significantly improves classifier performance over zero-shot by 3.8 percent, on average. All F1 gains for civil, uncivil and the overall model are statistically significant at the 0.01 level.\footnote{We run two-sample t-tests with both sets of 10x runs to determine statistical significance between two F1 scores. This applies to all results below where we mention a statistically significant difference.} Importantly, memory prompting improves on few-shot CoT significantly, especially for civil tweets, where the $F1$ score jumps to 0.820, a 9.19 percent gain. For uncivil tweets, the $F1$ score remains at 0.842, but for the overall model it increases to 0.832, an increase of 5.85 percent. These improvements are statistically significant and show that model memory can yield important performance gains, especially considering ceiling effects.\footnote{Most scores are around or above 0.8, leaving less room for improvement.} Gains for memory reinforcement are slightly more modest and statistically indistinguishable from those with the memory prompting approach. They are, however, equally significant on average when compared to both approaches without memory (zero- and few-shot), especially in the civil category and overall model performance. 

With Llama (Table 3b), the results show more marked improvements when using either the memory prompting or the memory reinforcement approaches. For civil tweets, memory prompting improves performance by 24.41 percent --0.639 vs. 0.795-- when compared to zero-shot with no memory, and 4.8 percent over few-shot CoT with no memory. For uncivil tweets, the gain is 5.3 percent over zero-shot. The $F1$ score is the same as with few-shot CoT, but overall model performance improves by 14.1 percent over zero-shot and 2.4 percent over few-shot CoT, both statistically significant differences. The largest performance gains, however, are in the memory reinforcement strategy with Llama. For civil tweets, the reinforcement $F1$ score stands at 0.830, 4.4 percent better than the memory prompting approach (0.795), 9.94 percent higher than few-shot CoT (0.755), and 29.89 percent higher than the zero-shot no-memory approach. With uncivil tweets, the gains are also significant. Reinforcement bests zero-shot no-memory by 9.19 percent (0.783 vs 0.855) and memory prompting by 3.26 percent (0.828 vs 0.855). Overall performance is also significantly improved using memory reinforcement, with gains of 18.42 percent when compared to the zero-shot no-memory approach (0.711 vs 0.842) and 4.2 percent when compared to the memory prompting strategy (0.811 vs 0.842). 

% Three additional findings merit mention. First, we test the degree to which we can attribute the raise in performance to memory. To this end, we estimate the running averages of F1 scores for few-shot CoT and no restart techniques, and plot them in Figure XX. What is readily apparent is that F1 scores in the no restart techniques improve as the model labels more instances. The same is not true for the few-shot CoT, where the evolution is flatter, as well as more variable. 

Two additional findings merit mention. First is the significant increase in recall scores for both LLMs, especially with the memory prompting approach. Differences are significant in both datasets but stronger with the nostalgia data. With GPT-4o, no-memory approaches lead to a large imbalance between false negatives and false positives. Zero-shot has an average of 13 for the former and 112 for the latter, while few-shot CoT has averages of 12 and 96, respectively. This improves significantly with memory prompting, which averages 30 false negatives and 70 false positives. With Llama, large imbalances are also reduced with memory prompting, which has 32 false negatives and 83 false positives. As a consequence, for both models, memory prompting (GPT-4o) and memory reinforcement (Llama) produce recall scores that are significantly higher than no-memory approaches. Notice that precision also increases for the uncivil category with memory prompting and memory reinforcement, due to fewer false positives. With the nostalgia data, recall scores are also significantly higher when using memory prompting and memory reinforcement, indicating a greater balance between false negatives and false positives.

Second, there are important differences across the two LLMs. On the one hand, while both memory prompting and memory reinforcement perform consistently better than either of the no-memory approaches, it is not always the case that reinforcement yields better results than memory prompting. In the nostalgia task, reinforcement outperforms all other approaches with both LLMs, but memory prompting is generally better than reinforcement with GPT-4o in the incivility task. We suspect this may be due to task complexity: classifying incivility might be easier than nostalgia, as evinced by the lower zero-shot $F1$ scores in the nostalgia task, so the reinforcement step may lead to greater improvements with more complex texts. 

Lastly, we want to note a few smaller findings that may be relevant to applied researchers. First is Llama's greater capacity to process larger amounts of tokens more rapidly and with less API errors and rate limitations. In our tests (see Table B1 in Appendix B), we find that Llama tends to run faster than GPT-4o and rarely yields an API error. Conversely, GPT-4o has limits on how many tokens per minute one can send to the API.\footnote{See Appendix B for a complete breakdown of these limits.} These limits may affect researchers who spend smaller amounts on the platform, especially with tasks that require sending long texts or conversation histories to the API. The reinforcement approach is such a task. Other important considerations for applied researchers are cost and time of running these models, which we report for all models in Appendix B. Llama is generally cheaper (or free) and faster than GPT-4o.\footnote{While Llama is open-source, a server is required to run the model through the API, which is the most convenient way to run it. Users can also look to run it for free either in a local computer or through institutional compute facilities.}

%% \footnote{API access for OpenAI models is based on a tier system set by level of spending on the platform. For tier 2 users (\$50 total spend), the limit is 450,000 tokens per minute (TPM). For tier 3 (\$100), the TPM is 800,000, for tier 4 (\$250) it is 2 million, and for tier 5 (\$1,000) it is 30 million.}
%% time of the runs -- table in App similar to JOP article

\vspace{1cm}

\section*{Implications and Conclusion}

\noindent In this article we address an important and understudied question in LLM annotation prompting: does model \textit{memory} affect classification performance? We test four approaches: zero-shot and few-shot CoT with no memory, the most common approaches currently in the literature; memory prompting, where we allow the model to have a memory of previous classifications; and memory reinforcement, where the model is told about its performance using a training set (20\% of the sample) in an interactive session, and then allowed to classify the rest of the sample. We test these approaches using OpenAI's GPT-4o and Meta's Llama 3.1 on two political science datasets (incivility on Twitter and political nostalgia). Our main finding is that memory matters: allowing the model to keep a history of previous interactions significantly affects future performance, as does reinforcing whether a subset of previous classifications was correct or not. With both GPT-4o and Llama, the common no-memory approaches perform significantly below the memory prompting and memory reinforcement approaches. 

Our findings have major implications for applied researchers using LLMs for annotation tasks, who may want to use either a memory prompting or memory reinforcement approach rather than the more common zero-shot or few-shot CoT no-memory strategies. More consistent and accurate annotations have a positive impact on downstream tasks, in particular when using the resulting data as a training set. Fine-tuned encoder-decoder models like RoBERTa may still outperform generative models in some text classification tasks (e.g., sentiment), while being considerably faster and cheaper \cite{bucher2024fine}. However, transformers-based models require high-quality annotated training sets for optimal performance \citep{timoneda2025}, which can be costly. With generative models, researchers can implement \textit{memory prompting} and \textit{memory reinforcement} to improve-task specific performance efficiently. 

Given the promising results, there are various lines of inquiry to further explore. The more balanced ratio between false positives and false negatives obtained through memory prompting and memory reinforcement suggests a more consistent reasoning from LLMs that avoids falling into repetitive labeling patterns (e.g., predicting the same label for most texts). Future research can explore the extent to which this is true for tasks with varying degree of complexity. Additionally, a model that relies on memory and information for prediction, is likely to be less reliant on the grammatical architecture of the prompt itself. This can have important implications in terms of replicability and consistency. 

\vspace{1cm}

\setstretch{1.5}

\bibliography{reference}
\bibliographystyle{apsr}
\nocite{}

\newpage

\includegraphics[scale=.75,page=1]{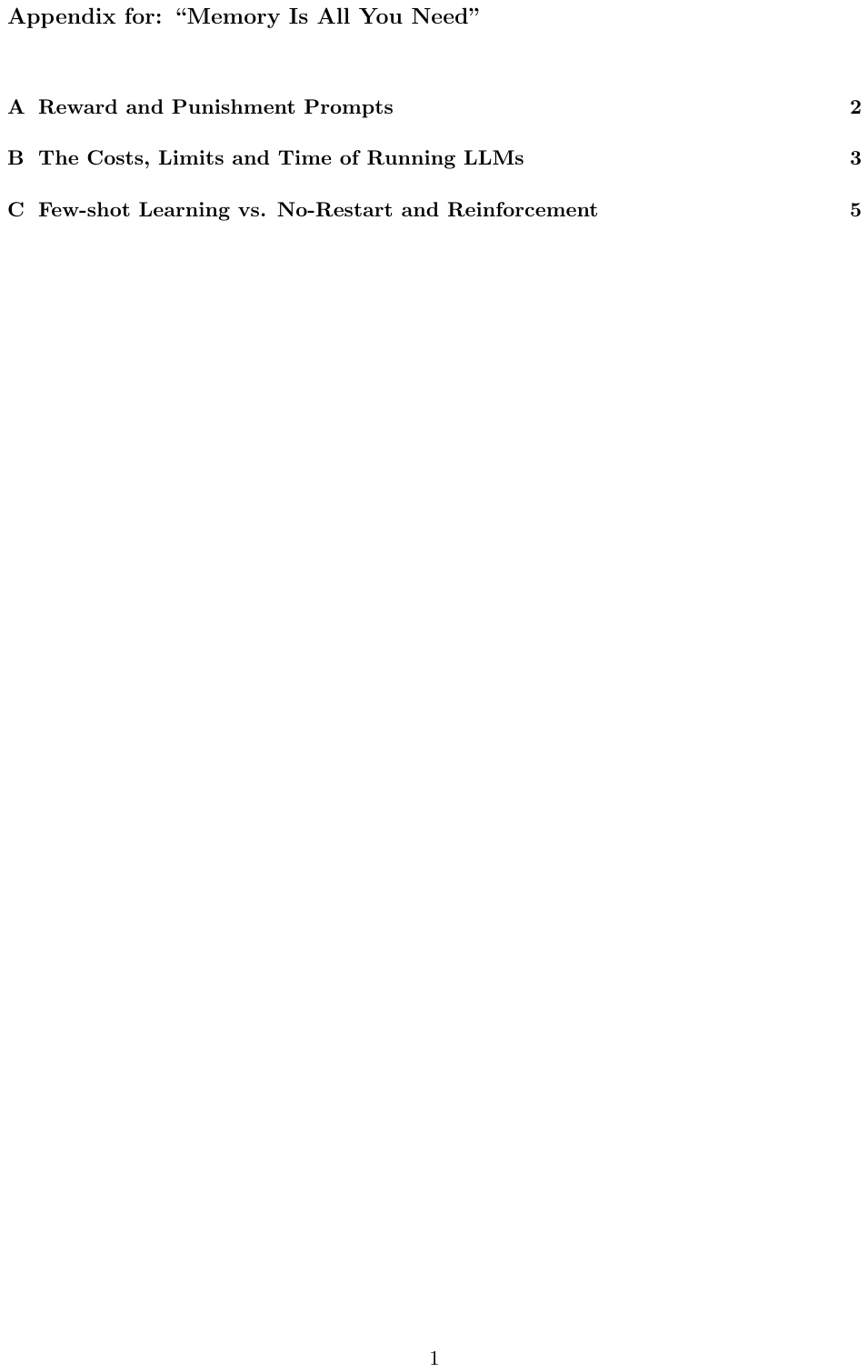}
\newpage

\includegraphics[scale=.75,page=2]{appendix.pdf}
\newpage

\includegraphics[scale=.75,page=3]{appendix.pdf}
\newpage

\includegraphics[scale=.75,page=4]{appendix.pdf}
\newpage

\includegraphics[scale=.75,page=5]{appendix.pdf}
\end{document}